\documentclass[10pt,twocolumn,letterpaper]{article}

\usepackage[pagenumbers]{cvpr} 

\newcommand{\todo}[1]{}
\newcommand{\TODO}[1]{\textbf{}}

\usepackage{cuted}
\usepackage{multirow}
\definecolor{cvprblue}{rgb}{0.21,0.49,0.74}
\usepackage[pagebackref,breaklinks,colorlinks,allcolors=cvprblue]{hyperref}

\def\confName{CVPR}
\def\confYear{2027}

\newcommand{\acronym}{SWAM\xspace}

\title{\acronym: State-aware World Action Model for Multi-Agent 3D Environments}

\author{
Jenna Kang\\
New York University\\
{\tt\small jennakang@nyu.edu}
\and
Bingxuan Li\\
New York University\\
{\tt\small bingxuan.li@nyu.edu}
\and
Colin Groth\\
New York University\\
{\tt\small cg4932@nyu.edu}
\and
Anatolii Evdokimov\\
New York University\\
{\tt\small tolya.e@nyu.edu}
\and
Sai Qian Zhang\\
New York University\\
{\tt\small sai.zhang@nyu.edu}
\and
Qi Sun\\
New York University\\
{\tt\small qisun@nyu.edu}
}

\begin{document}
\maketitle
\begin{strip}  
\centering
  \includegraphics[width=\linewidth]{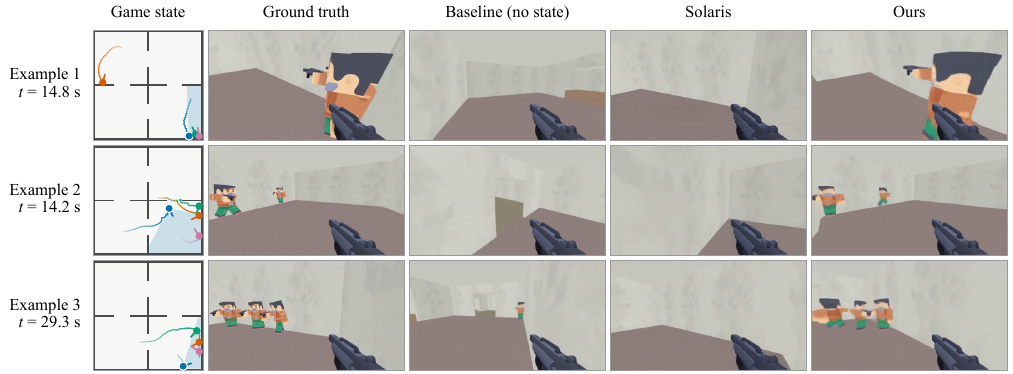}
  \captionof{figure}{\textbf{Illustration of our output and qualitative comparisons with comparison methods.} Each row is one test clip; all columns show the same frame. The left panel is the recorded game state (top-down; the camera player's field of view is shaded). Our state-conditioned model renders the other players accurately, while the baseline without state and Solaris fail to reconstruct the player correctly.}
  \captionsetup{hypcap=true}

  \label{fig:rendering_quality}
\end{strip}
\begin{abstract}
Interactive world models generate virtual environments autoregressively from player actions and visual history. However, visual history alone cannot reliably capture the evolving states of independently acting non-player characters (NPCs) in real games, particularly when they are outside the camera view. Recent multi-agent world models address this challenge through joint multi-view generation or persistent shared state, but often rely on external observations, like maps or scene geometry, or predefined state transitions. We present \acronym, a State-aware World Action Model that separates agent-state evolution from autoregressive visual generation in interactive 3D environments. A lightweight state model propagates character dynamics from actions without consuming generated frames, combining motion priors with learned scene-dependent corrections. The predicted states then guide an autoregressive diffusion transformer alongside visual history through a novel conditioning scheme. Rather than requiring an explicit scene geometry map or visibility tests, the conditioning indicates where characters may appear, allowing the diffusion model to learn their appearance and occlusion. This design enables NPCs to evolve independently of visual generation while maintaining consistent observations across visibility changes, interactions, and respawns. Evaluations in a custom Godot-based first-person shooter game demonstrate improvements in visual fidelity, NPC consistency, and long-horizon stability over existing approaches. 
\end{abstract}

%
%
%
%
\section{Introduction}
\label{sec:intro}
%
Most existing generative world models rely on autoregressive prediction, where previously generated frames provide the context for generating future observations. While effective at maintaining short-term visual continuity, this paradigm faces challenges in multi-agent environments, where independently acting players and non-player characters (NPCs) move, interact, and change states inside and outside the ego player's field of view. Such changes cannot be reliably inferred from the ego player's action and visual history alone, leading to inconsistencies when characters reappear after prolonged absence or occlusion. Recent work has addressed multi-agent consistency through two broad directions. The first jointly models multiple player perspectives, as in Solaris \cite{savva2026solaris} and MultiWorld \cite{wu2026multiworld}, enabling synchronized multi-view generation through coordinated training and shared multi-agent representations. 
The second introduces persistent state representations to decouple world evolution from visual generation, like in MultiGen \cite{po2026multigen}, where an explicit map of the game is used to express game state and perform visibility tests. 
Despite these advances, existing approaches employ different forms of explicit state representations, distributed coordination, or scene-level structure to maintain consistency. This motivates a complementary approach that propagates the dynamics of independently acting agents while retaining autoregressive visual history for high-fidelity generation.

By asking whether an explicit scene representation is necessary for consistent multiplayer generation, we present \acronym, a State-aware World Action Model that combines action-driven agent-state propagation with history-conditioned visual generation. \acronym maintains a lightweight representation of character states that evolves independently of the generated video. A small \emph{state model} combines simple motion priors with learned corrections to predict character positions, orientations, and interaction states directly from actions, without rendered observations or external scene geometry. 
Consequently, characters can continue evolving even when they remain outside the ego player's view, and their states can be shared across independently generated viewpoints.
A diffusion \emph{world model} then generates the ego frames from the state using a novel \emph{FOV conditioning}: predicted character states are projected into spatial indicators that encode their possible appearance in a player's ego frame. Unlike conventional geometry-based rendering, these indicator masks do not require explicit visibility tests. The diffusion world model learns to use the mask together with previously generated visual latents to synthesize temporally coherent observations and learn scene-dependent player occlusion. In this way, visual history preserves appearance and scene continuity, while explicit agent states provide information about character dynamics that cannot be recovered from visual history alone.


Removing the explicit map and explicit visibility maps changes what the system can do. The approach applies to full 3D environments with free 6-DoF camera motion, which a 2D layout does not capture. Characters are simulated by the state model alone and incur no generation cost, while each human player only needs to run one instance of the same lightweight world model. Because character identity and location come from the propagated state rather than from past frames, we show that characters remain consistent across occlusions, long absences from the view, and respawns.
We evaluate the approach in a custom-designed 3D first-person shooter. Our model sustains consistent gameplay even over long-horizon rollouts of 10 minutes with the player's own actions as the only external input where comparison method fail to render other characters consistently. In summary, our contributions include:
\begin{itemize}
    \item A factorized multiplayer world model that requires no explicit scene representation: the actions of the characters are the only explicit input, and scene geometry is learned implicitly.
    \item A lightweight state model that propagates the state of all characters from actions alone, without rendered observations, so that bots and unseen players are simulated at negligible cost.
    \item An FOV conditioning scheme based on box indicators whose size does not grow with the number of characters and which keeps other players and NPCs consistent.
\end{itemize}
\section{Related Work}
\label{sec:related_work}

\subsection{Action-Conditioned Interactive World Models}
Latent world models learn predictive dynamics for planning and control~\cite{hafner2019planet,hafner2025dreamer}. More recently, {Genie}~\cite{bruce2024genie} learns controllable environments from video, and {DIAMOND}~\cite{alonso2024diamond} and {GameNGen}~\cite{valevski2025gamengen} demonstrate diffusion-based game simulation. {GameFactory}~\cite{yu2025gamefactory} extends generation to novel game scenes, while {Vid2World}~\cite{huang2026vid2world} adapts video diffusion to causal, action-conditioned prediction. Real-time and streaming generation has been explored by {Matrix-Game 2.0}~\cite{he2025matrixgame2}, {Self Forcing}~\cite{huang2025selfforcing}, and {WorldPlay}~\cite{sun2026worldplay}. In first-person 3D games, {WorldCam}~\cite{nam2026worldcam} uses camera poses to connect action control with geometrical consistency. However, these advances primarily address the controllability, visual quality, or geometric consistency of generated views; plausible ego-view prediction alone does not ensure that independently acting characters continue to evolve correctly while unseen. This motivates explicitly propagating agent dynamics outside the visual generation loop.

\subsection{Long-Horizon Memory and Multi-Agent Generation}
Long-range coherence is a central limitation of autoregressive visual simulation. {Long-Context State-Space Video World Models}~\cite{po2025longcontext} extend temporal memory through state-space modeling; {ReWorld}~\cite{chen2026reworld} uses pose-indexed retrieval and bounded memory, while {ActWorld}~\cite{xiong2026actworld} retains action-relevant object and interaction events. {PERSIST}~\cite{garcin2026persist} instead maintains a latent 3D scene to improve spatial consistency. These methods address remembering previously observed locations or events, whereas multi-agent gameplay additionally requires predicting changes to entities that are not currently observed. {WRBench}~\cite{lu2026persistent} directly studies failures in persistent state evolution. {World Observer}~\cite{choi2026worldobserver} maintains out-of-view dynamics by jointly generating an actor view and additional observer views, an alternative that incurs visual generation for the observer streams.

Multi-agent world models increasingly coordinate multiple viewpoints and actions. {Solaris}~\cite{savva2026solaris} generates synchronized player views in Minecraft; {MultiWorld}~\cite{wu2026multiworld} incorporates global state into multi-agent video prediction; and {$\gamma$-World}~\cite{liu2026gammaworld} develops permutation-aware mechanisms for multi-agent generation. {Prisma-World}~\cite{sun2026prismaworld} couples viewpoints through geometry-aware generation, {PAWA}~\cite{yang2026pawa} models panoramic and first-person streams, and {ME-World}~\cite{chung2026meworld} jointly denoises egocentric videos conditioned on fine-grained actions and shared environment memory. {WorldWeaver}~\cite{mo2026worldweaver} augments multi-agent generation with persistent world-state registers. Although these approaches improve multi-view consistency, jointly generating player views incurs visual synthesis for streams that need not be displayed, while visual memory alone cannot determine unobserved actions and state changes. This motivates a world model that maintains NPC and off-screen player states without rendering their views.

\begin{figure}[t]   
  \centering
  \includegraphics[width=\linewidth]{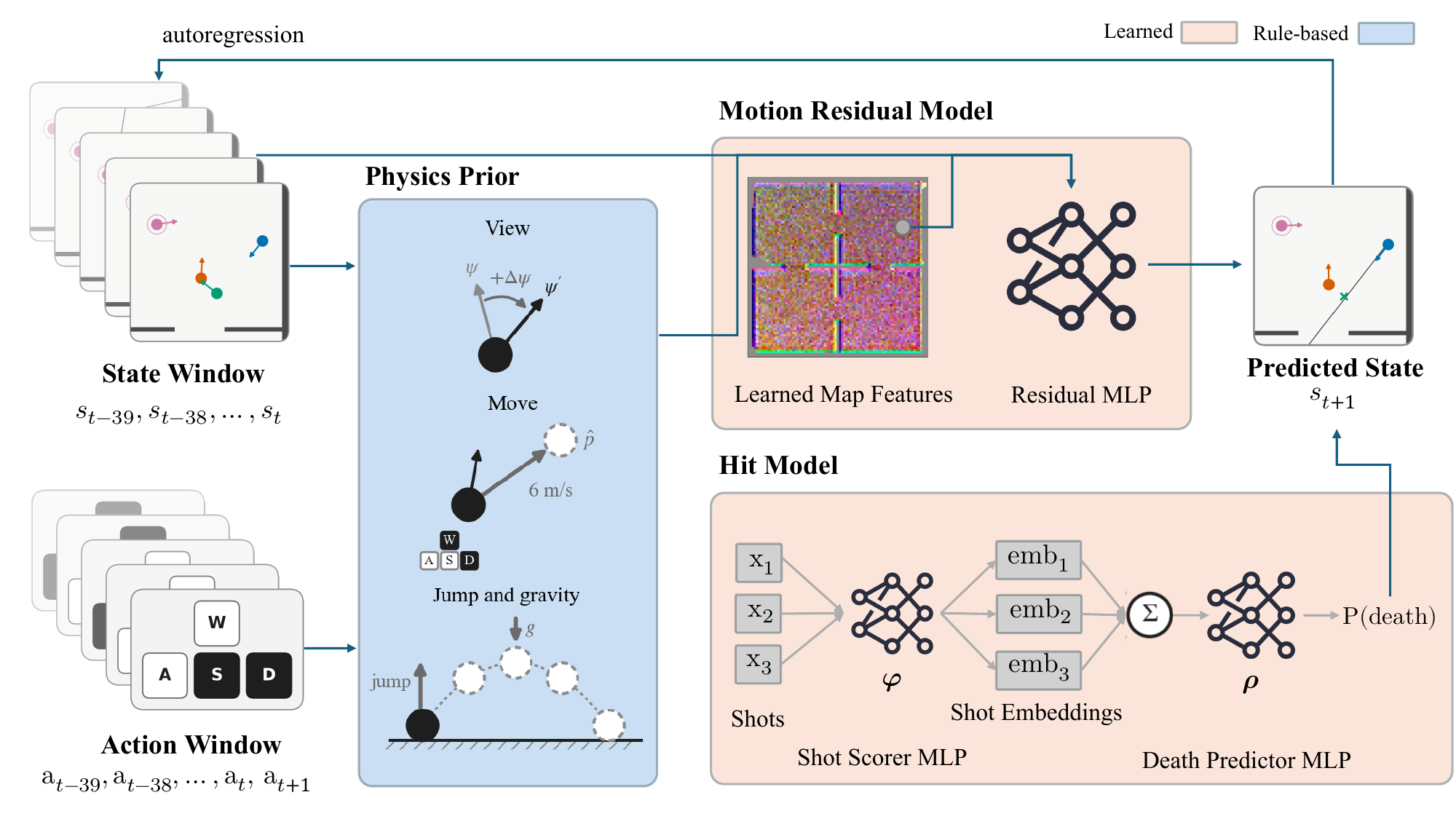}
\caption{\textbf{Overview of the state model.} The model predicts every player's next state $s_{t+1}$ from the last $W{=}40$ states and actions. Rules are shown in blue and learned components in orange.}
  \label{fig:state_model}
\end{figure}

\subsection{Shared State and Spatial Conditioning}
Explicit or factored state is increasingly used to separate world evolution from appearance generation. {MultiGen}~\cite{po2026multigen} factorizes multiplayer diffusion game engines into memory, observation, and dynamics components, using a specified 2D level map and observation-derived features. {WorldCast}~\cite{ye2026worldcast} supports distributed players with camera-aligned, fixed-resolution state conditioning and estimates geometry-related signals from generated observations. {MASS}~\cite{cai2026mass} uses authoritative shared state to drive action-dependent updates separately from visual synthesis. Particularly close to our approach, {Programmable World Model}~\cite{huang2026programmable} evolves a persistent global entity state through executable, user-specified rules and projects state-augmented 3D oriented boxes into pixel-aligned video conditions. Thus, neither separating state updates from rendering nor using projected boxes is new in isolation. Instead, we target a first-person 3D multiplayer setting with action-driven state propagation: motion priors and learned residuals update agent states without consuming generated observations, and fixed-resolution field-of-view box indicators mark possible agent locations without performing geometric visibility tests.
Existing approaches variously depend on supplied level layouts, rendering-derived state updates, or externally specified entity dynamics; they leave open how to combine low-cost autonomous agent simulation with occlusion-aware ego-view generation in a 3D multiplayer game. SWAM addresses this gap by learning the scene-dependent corrections to agent dynamics and letting the video generator resolve visibility from simple state-projected box indicators.

\section{Method}
\label{sec:met}

\begin{figure*}[t]
  \centering
   \includegraphics[width=\linewidth]{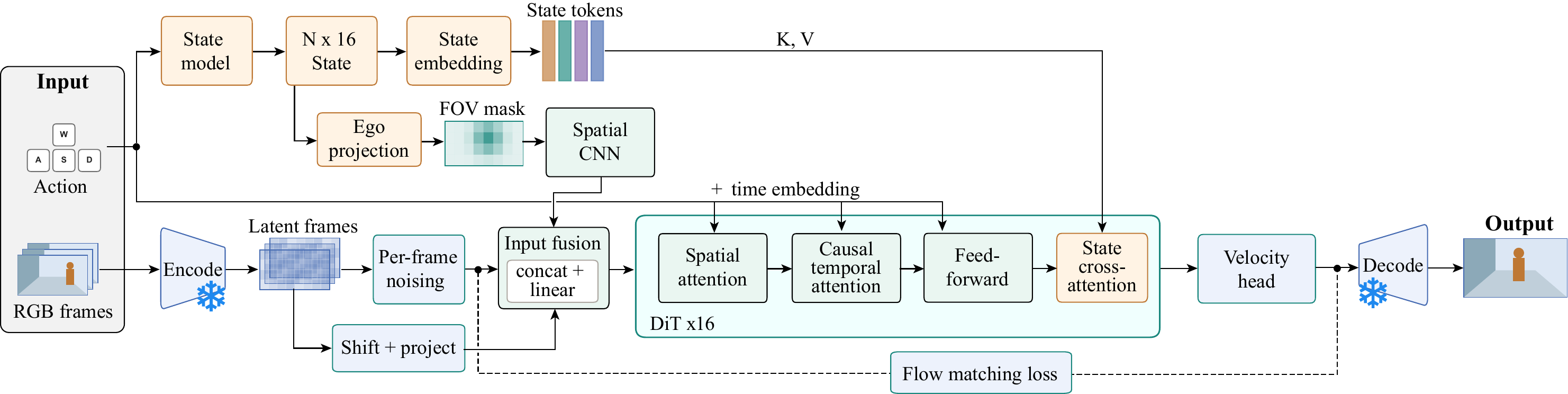}
   \caption{\textbf{Overview of our method.} Dense player masks provide spatial conditioning, while state cross-attention injects player information into the video diffusion model.}
   \label{fig:pipe}
\end{figure*}

%
Our system couples a state model that predicts multiplayer dynamics with a world model that renders ego-view video from actions and shared state (see \cref{fig:pipe}). We first train a video autoencoder, then freeze it and train the world model in its latent space; the state model is trained separately.

\subsection{State Model}
We model the game as a discrete-time dynamical system over the explicit state of all players. At each frame $t$ (20\,Hz), player $i \in \{1,\dots,4\}$ has state $s^i_t = (\mathbf{p}^i_t, \mathbf{v}^i_t, \psi^i_t, \theta^i_t, d^i_t, g^i_t)$: position, velocity, yaw, pitch, and the dead and grounded flags. Its action $a^i_t$ consists of six binary keys (move, fire, jump) and the look change $(\Delta\psi, \Delta\theta)$. Given the last $W=40$ frames of all players and their next actions, the model predicts the next state of every player,
\begin{equation}
  s_{t+1} = f_\phi\bigl(s_{t-W+1:t},\ a_{t-W+2:t+1}\bigr),
\end{equation}
and is rolled out autoregressively by appending each prediction to the window.

To facilitate accurate state prediction, we combine deterministic game rules with learned residual dynamics (see supplementary). A physics prior gives the next velocity $\hat{\mathbf{v}}$ and position $\hat{\mathbf{p}}$ of a living player from its keys, gravity and jumps. It is exact in open space but wrong at walls, on landings, and on some decision frames of the bots, so an MLP predicts residuals:
\begin{equation}
  \mathbf{v}_{t+1} = \hat{\mathbf{v}} + \boldsymbol{\sigma}_v \odot \Delta\mathbf{v}, \qquad
  \mathbf{p}_{t+1} = \hat{\mathbf{p}} + \boldsymbol{\sigma}_p \odot \Delta\mathbf{p},
\end{equation}

where $\boldsymbol{\sigma}_v$ and $\boldsymbol{\sigma}_p$ scale residuals using training statistics. The MLP also predicts the grounded flag and takes three recent states and actions, the physics prior, other players, and a learned spatial feature grid. Its zero-initialized output recovers the prior before training.
Kills depend on individual shots rather than accumulated damage. We therefore encode each shot separately and aggregate shots fired at a player over the last 10 frames to predict death:
\begin{equation}
  P(\text{dies}) = \sigma\Bigl(\rho\Bigl(\textstyle\sum_{s} \varphi(x_s)\Bigr)\Bigr),
\end{equation}

Here, $x_s$ encodes aim, hitbox-relative trajectory, and weapon readiness; $\varphi$ and $\rho$ are MLPs and $\sigma$ is the logistic function. At inference, deaths are sampled from this probability; dead players remain frozen for 30 frames before respawning uniformly at one of eight spawn points. See \cref{fig:state_model} for the architecture.

We train for 20k steps, using teacher-forced Smooth-L1 loss on normalized motion residuals and binary cross-entropy on grounded and death flags. After 12k steps, we add position and velocity losses back-propagated through autoregressive rollouts. Ground-truth deaths and spawn points isolate motion learning in both phases.

\subsection{World Model}
\begin{figure*}[t]   
  \centering
  \includegraphics[width=\linewidth]{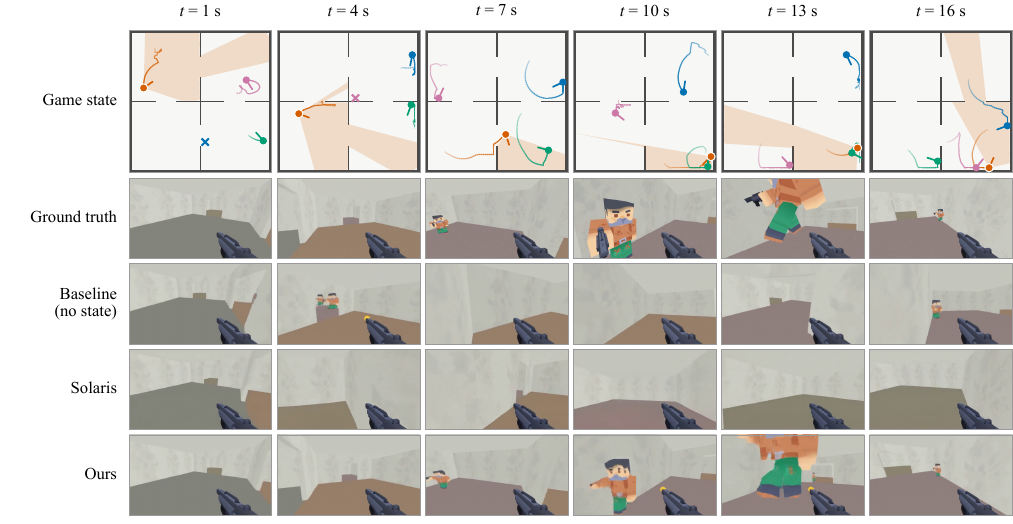}
  \caption{\textbf{Frame progression over time.} Recorded game state at each time step, with the camera player's field of view shaded (top row). Ground truth is rendered in-game. Ours preserves visible characters in the FOV, while in the comparison methods they disappear.}
  \label{fig:temporal}
\end{figure*}

The world model conditions a latent diffusion transformer on ego actions, visual history, global player-state tokens, and a FOV-projected dense player mask.

\paragraph{Video autoencoder training.}
A representation autoencoder (RAE) maps video to standardized latents $Z\in\mathbb{R}^{L\times H\times W\times C_z}$, with $L=T/2$ and $(H,W)=(9,16)$. Latent $k$ represents RGB frames $(2k,2k+1)$ and receives state at $2k$ and incoming ego actions $(a^e_{2k-1},a^e_{2k})$.
We train a DINO-based video encoder $\mathcal E_{\rm vid}$ and transformer decoder $\mathcal D_{\rm vid}$ to reconstruct $\hat X=\mathcal D_{\rm vid}(\mathcal E_{\rm vid}(X))$ using pixel $\ell_1$, LPIPS, and DINO-feature consistency losses. The pretrained DINO backbone is fixed while the spatiotemporal bottleneck and decoder are trained. We then freeze the entire codec for latent world-model training.

\paragraph{Player-state conditioning.}
The world model is conditioned on the same per-player state $s^i_t$ that the state model predicts. Each player $i$ is encoded as
\begin{equation}
\begin{aligned}
\mathbf E^i_t &= \operatorname{MLP}_{\rm state}(s^i_t)+\mathbf e_i,
\end{aligned}
\label{eq:wm-state}
\end{equation}
 A two-layer SiLU MLP and learned slot embedding $\mathbf e_i$ produce four identity-preserving tokens of width $D=2048$, including off-screen players.

\paragraph{Projected FOV mask.}

We project $1.8\,\mathrm{m}\times0.7\,\mathrm{m}$ upright player proxies through the ego camera and place unit-peak Gaussians with a standard deviation of one latent cell at their clipped box centers on the latent grid. Self, dead, behind-camera, and out-of-frustum targets contribute zero, as do all maps for a dead ego. These maps encode geometric placement without testing scene occlusion.

At each cell, we rank players by their Gaussian contributions exponentially discounted by normalized camera-forward depth, retaining the top two players. Depth is log-scaled and normalized to $[0,1]$, with $100\,\mathrm{m}$ mapped to one.

The dense mask $F^e\in\mathbb{R}^{L\times8\times H\times W}$ contains coverage, depth, relative-yaw sine and cosine, and four player-slot maps. Coverage is the clipped sum of retained Gaussians; depth and orientation are their unnormalized weighted sums. Slot maps retain each selected player's Gaussian contribution.

\paragraph{Conditioned visual transformer.}
A spatial encoder $\mathcal E_F$ maps the mask to 128 channels using a $3\times3$ convolution, SiLU, and a zero-initialized $1\times1$ output convolution. Its features are concatenated with each noisy latent before projection to the transformer width:
\begin{equation}
\begin{aligned}
R_k&=\mathcal E_F(F^e_k),\\
H^0_k&=\mathcal P_{\rm in}([Z^{\tau_k}_k;R_k])+
        \mathcal P_{\rm past}(P_k).
\end{aligned}
\label{eq:wm-input}
\end{equation}

The per-cell affine projections fuse the noisy current latent with the clean previous latent $P_k=Z_{k-1}$, obtained from the frozen encoder during training and from generated predictions at inference. A learned grid initializes the first step. Pooled ego-action and flow-time embeddings provide adaptive LayerNorm conditioning.

The backbone has 16 blocks of width 2048, each applying spatial self-attention, optional causal temporal attention, a SwiGLU feed-forward network, and player-state cross-attention, with residual connections. Self-attention uses 16 query and four KV heads, with rotary position embeddings; temporal attention appears in blocks 1, 5, 9, 13, and 16.

In state cross-attention, each frame's 144 visual tokens query its four player tokens as keys and values, using separate layer normalizations and 16 heads. This complements the local mask with global access to player identity, motion, and status. A final affine head predicts latent velocities $\hat V\in\mathbb{R}^{L\times H\times W\times C_z}$.

\paragraph{Latent world-model training.}

With the codec frozen, we independently sample $\tau_k\sim\mathcal U(0,1)$ and $\epsilon_k\sim\mathcal N(0,I)$ for each latent frame:
\begin{equation}
Z_k^{\tau_k}=\tau_k Z_k+(1-\tau_k)\epsilon_k,
\qquad V_k=Z_k-\epsilon_k.
\label{eq:wm-flow-target}
\end{equation}

We train the transformer and conditioning modules with flow matching, $\mathcal L_{\rm WM}=\mathbb E[\|\hat V_k-V_k\|_2^2]$, using ground-truth states and derived masks. Per-frame noising and causal attention implement diffusion forcing \cite{diff_forcing}, supplemented by clean-past fusion (\cref{eq:wm-input}). The codec and explicit state model receive no updates from this loss.

The frozen encoder initializes visual context. We generate each new latent from Gaussian noise by Euler updates, $Z^{\tau+\Delta\tau}=Z^\tau+\Delta\tau\hat V$, and reuse the result as clean-past context. The frozen decoder converts generated latents to RGB after reversing standardization.

\section{Experiments}\label{sec:exp}\label{sec:experiments}

\begin{table*}[t]
    \centering
    \caption{\textbf{Main quantitative results.}
    We compare our method to a baseline without explicit state conditioning and a comparison method that renders all player frames in one model (Solaris). Best results are in \textbf{bold}. MILO values are scaled by $10^{3}$.
    FVD subscripts denote clip length in frames. Player \textit{recall} and \textit{precision} use FOV-filtered YOLO detections (FOV overlap $\geq 0.2$), matched at
    IoU$\geq0.2$. \textit{Time per frame} is causal end-to-end time per generated RGB view. \textit{Params} denotes total model parameters, including the encoders.}
    \label{tab:comparison_horizons}
    \label{tab:comparison}
    \resizebox{\textwidth}{!}{%
    \begin{tabular}{llcccccccccc}
        \toprule
        Horizon & Method
        & LPIPS$\downarrow$ & MILO$\downarrow$ & SSIM$\uparrow$ & PSNR$\uparrow$
        & FVD$_{16}\downarrow$ & FVD$_{32}\downarrow$
        & Recall [\%]$\uparrow$ & Precision [\%]$\uparrow$
        & Time / Frame [ms]$\downarrow$ & Params [B]$\downarrow$ \\
        \midrule
        \multirow{3}{*}{10\,sec.}
        & Baseline (no state)
        & 0.305 & 11.74 & 0.897 & 22.65 & 132.3 & 243.9
        & 45.7 & 52.2 & 160 & \textbf{2.09} \\
        & Solaris \cite{savva2026solaris}
        & 0.317 & 10.71 & 0.885 & 21.99 & 192.9 & 348.3
        & 39.4 & 57.7 & 960 & 2.58 \\
        & Ours 
        & \textbf{0.223} & \textbf{8.33} & \textbf{0.913} & \textbf{25.56}
        & \textbf{86.9} & \textbf{126.6}
        & \textbf{74.5} & \textbf{69.8} & \textbf{158} & 2.36 \\
        \midrule
        \multirow{3}{*}{30\,sec.}
        & Baseline (no state)
        & 0.344 & 13.05 & 0.891 & 21.06 & 160.2 & 299.8
        & 28.3 & 40.3 & 160 & \textbf{2.09} \\ 
        & Solaris \cite{savva2026solaris} 
        & 0.371 & 12.32 & 0.870 & 20.33 & 262.0 & 440.2
        &  22.6 & 47.3 & 960 & 2.58 \\
        & Ours 
        & \textbf{0.249} & \textbf{9.53} & \textbf{0.904} & \textbf{23.90}
        & \textbf{101.3} & \textbf{137.5}
        & \textbf{71.1} & \textbf{70.1} & \textbf{158} & 2.36 \\
        \midrule
        \multirow{3}{*}{10 min.}
        & Baseline (no state)
        & 0.409 & 14.87 & 0.882 & 19.04 & 212.8 & 336.8
        & 2.0 & 3.0 & 160 & \textbf{2.09} \\
        & Solaris \cite{savva2026solaris}
        & 0.439 & 15.01 & 0.813 & 18.48 & 600.4 & 745.3
        & 1.1 & 6.9 & 960 & 2.58 \\
        & Ours 
        & \textbf{0.336} & \textbf{12.53} & \textbf{0.885} & \textbf{20.47}
        & \textbf{160.6} & \textbf{188.9}
        & \textbf{57.2} & \textbf{55.0} & \textbf{158} & 2.36 \\
        \bottomrule
    \end{tabular}}
\end{table*}

\subsection{Experimental Setup}
\label{sec:setup}

\paragraph{Testbed and Dataset.}
We collect gameplay data in a custom Godot first-person shooter \cite{godotengine}, which records every player's actions and GT state. The layout motivates frequent field-of-view changes and wall occlusions (cf. \cref{fig:temporal}).
For data collection, we train four bots that share a PPO-trained MLP policy \cite{schulman2017PPO}. Details on the RL bot training are in the supplementary.
The collected dataset contains $40$h of four-player gameplay at $512\times288$ and $20$fps. We split whole matches into $80/20$ training/testing and use the same split for both models.

\paragraph{Comparison Methods.}
We compare with (i)~\emph{Baseline (no state)}, an action-conditioned world model \cite{hu2026multiplayerinteractiveworldmodelsMIRA} receiving all players' actions without explicit state, and (ii)~\emph{Solaris} \cite{savva2026solaris}, which jointly generates all player views. Starting from Solaris's official two-player Minecraft checkpoint, we extend its player embeddings from two to four slots and fine-tune all 1.8B parameters. Solaris benefits from large-scale Minecraft pretraining, whereas our model and the no-state baseline are trained from scratch. All methods receive the same GT visual context.
We exclude MultiGen~\cite{po2026multigen} because its top-down-map depth conditioning is not directly applicable to our free-pitch and jumping setting.

\paragraph{Metrics.}
We report LPIPS \cite{zhang2018lpips}, MILO \cite{milo_metric}, SSIM \cite{ssim}, PSNR, and FVD \cite{fvd} over 16- and 32-frame windows, as well as inference time per frame and memory per client.

To measure character accuracy beyond full-frame metrics, we fine-tune COCO-pretrained YOLO11m \cite{yolo} detectors on several hundred paired generated and GT frames per model, with additional auto-labeled GT data for the skin detector. Detector-proposed annotations are refined by Claude Opus 5.5 \cite{anthropic2026opus55systemcard}, with author spot-checks. On held-out detector evaluation data, generated/GT mAP@0.5 reaches 0.91/0.96 (single-skin), 0.74/0.97 (varied-skins), and 0.99/0.96 (Solaris single-skin).

We retain GT player boxes overlapping recorded FOV boxes at IoU $\geq 0.2$, then match generated and eligible GT detections one-to-one at the same threshold. Recall measures the fraction of eligible GT characters rendered at the correct location, while precision measures the fraction of generated characters matched to eligible GT characters. Both are averaged per rollout and then across rollouts. In the 1 vs.\ 1 setting, only the unmasked target is eligible; detections matching masked-player GT boxes are excluded.

We evaluate autoregressive rollouts initialized with 40 GT context frames (2\,s at 20\,fps), conditioned on recorded actions and, for state-conditioned evaluations, player states. The test set contains 12 episodes with four POVs each. We evaluate 960 30\,s rollouts ($T=600$ generated frames), using 20 tail-aligned windows per POV at 600-frame intervals, with a 40-frame overlap between the final two windows. We report both 10\,s ($T=200$) and 30\,s horizons. Additionally, we evaluate 48 continuous full-episode rollouts, each generating $T=11{,}960$ frames (598\,s, approximately 10\,min). Each rollout is sampled once.

\paragraph{Implementation Details.}

Our world model has 1.46B parameters, excluding the frozen 900M-parameter RAE. We train it for 250k steps with AdamW, learning rate $10^{-4}$, batch size 1, weight decay 0.1, momentum $(0.9,0.99)$, 1k warmup steps, and EMA decay 0.9999. State-conditioned variants use GT states during training.

The state model has 300K parameters and is trained for 20k steps (approximately 20 epochs) with AdamW, learning rate $10^{-3}$, weight decay $10^{-4}$ ($10^{-3}$ for the hit model), batch size 512, 500 warmup steps followed by cosine decay, and gradient clipping at 1.0. The first 12k steps use teacher-forced predictions; the remaining 8k additionally use a rollout loss on 128 windows of 24 frames (1.2\,s) per step, as described in \cref{sec:met}.
Sampling uses 10 Euler steps with a linear--quadratic time schedule. All experiments are run on a single NVIDIA H200 GPU.

\subsection{Quantitative Results}\label{sec:quant_results}
\begin{figure*}[t]   
  \centering
  \includegraphics[width=\linewidth]{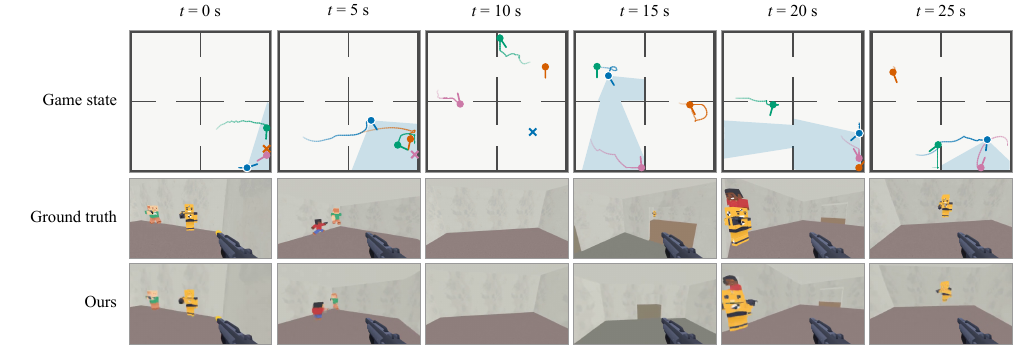}
  \caption{\textbf{Varied player skins.} Engine renderings and our model's outputs on a progressing test clip. The top row visualizes the player positions and the main player's field of view (shading).}
  \label{fig:temporal_skins}
\end{figure*}
\begin{table}[t]
  \centering
  \caption{\textbf{Rollout error of the state model.} The model is rolled out autoregressively
    with actions generated by the RL agent. The three columns report different rollout horizons (10\,sec., 30\,sec, 10\,min.). Mean $\pm$ std over 3 seeds on 11 test episodes with 32 random starts per test episode.}
  \label{tab:rollout-error}
      \resizebox{0.49\textwidth}{!}{%
  \begin{tabular}{lccc}
    \toprule
    & 10\,s & 30\,s & 10\,min (full episode) \\
    \midrule
    Position (m)              & $0.396 \pm 0.017$ & $0.518 \pm 0.091$ & $0.717 \pm 0.254$  \\
    Velocity (m/s)            & $0.418 \pm 0.010$ & $0.460 \pm 0.021$ & $0.463 \pm 0.017$ \\
    View (deg)                & $1.1 \times 10^{-4}$ & $1.6 \times 10^{-4}$ & $3.3 \times 10^{-4}$ \\
    Grounded flag (\% frames) & $0.66 \pm 0.04$ & $0.99 \pm 0.23$ & $0.72 \pm 0.04$ \\
    \bottomrule
  \end{tabular}}
\end{table}

Our main results are reported in \cref{tab:comparison}. To enable full reference visual quality assessment, we run the metrics on model output generated with game engine player states. Otherwise, even small drifts in the predicted states would misalign the generated frames with the reference video and distort the metrics. End-to-end results and generated videos with predicted states from our state model are analyzed and can be found in the supplementary material.
Rollout errors of the state model are about 0.7\,m in a 10-minute rollout (see \cref{tab:rollout-error}) for our 20\,m$\times$20\,m scene.
Our method achieves the best overall visual quality by every metric. The improvements are most pronounced for the perceptual metrics LPIPS and MILO, which align more closely with human perception. Compared to the baseline without state, our method reduces LPIPS by 28\% and MILO by 27\% in 30\,s rollouts. The FVD scores indicate temporally stable results.

Based on the YOLO detections, our model accurately and consistently renders other players. In the 30\,s rollouts, our method on average achieves over 70\% recall while the baseline and Solaris are only able to render characters correctly in 22--28\% of frames.  

Although our conditioning does not explicitly encode occlusion, the dense player mask and FOV-guided cross-attention models leave 95.24\% and 95.28\% of detector-estimated non-visible player instances unrendered, respectively, in 30\,s rollouts. Together with the high visible-player recall, these results suggest that the models learn visibility-dependent rendering.

Additionally, since our method does not process different player views simultaneously, like Solaris, it runs faster on a model of similar size.

\subsection{Qualitative Results}
\label{sec:qualitative}

\cref{fig:rendering_quality} compares the methods on several test recordings, and~\cref{fig:temporal} shows how the generated frames evolve over the course of a rollout. Please also refer to the supplementary video. 
%
All methods produce temporally stable videos with largely consistent geometry: the scene does not jump or change abruptly between frames. Consistency between frames, however, does not imply correctness. Solaris maps actions to the world incorrectly and inconsistently, so that its frames deviate significantly from the reference. Its rollouts are plausible as videos, since autoregression keeps consecutive frames coherent, but they do not follow the actual level geometry, including camera rotation errors, missing doors, and changed appearance of the rooms. Our method, in contrast, renders the scene both consistently and correctly. In the few cases where the geometry drifts during long rollouts, we observe that the model has learned to use the reset at death and respawn to fully recover the correct view of the scene.

The comparison methods fail to render other players correctly even in short rollouts. Our method displays them consistently and accurately, including difficult situations in which two players occlude each other (see the supplementary video). The result is notable because the FOV boxes are a trivial projection of the state into the view of the player and carry no information about actual visibility. The model nevertheless learns to resolve the ambiguity and renders each player as visible, partially hidden, or fully occluded as the scene requires.

\subsection{Ablations}


\paragraph{Positional and State Conditioning.}
We compare position-only and full-state conditioning with two forms of FOV conditioning: attention routing and dense player masks (\cref{tab:ablations}). Position conditioning improves visual quality over the no-state baseline, while full-state conditioning further stabilizes background geometry. However, state alone does not reliably place players at the correct image locations.

FOV-guided cross-attention adds projected player maps to the visual-to-state attention logits, scaled by a learned coefficient per layer, biasing spatial tokens toward geometrically relevant player states. Compared with full-state conditioning, routing increases character recall from 28.4\% to 66.5\% and precision from 38.6\% to 66.2\%. Dense-mask conditioning instead injects projected coverage, depth, orientation, and identity features into the visual input. It performs best across all reported metrics, reaching 71.1\% recall and 70.1\% precision and reducing FVD$_{32}$ from 152.9 to 137.5 relative to routing.

Visual inspection supports these differences: position-only conditioning can produce incorrect doors and camera rotations, while full state preserves scene geometry but still misplaces players. Both FOV variants improve character localization, with the dense mask providing further gains through richer spatial conditioning.

\paragraph{Varied NPC skins.}
In the main experiments, all players share the same appearance (skin). To show that our model is not tied to a single appearance, we record a new dataset of 20\,h of gameplay in which the four characters have different skins (shown in the supplementary material) and retrain the world model from scratch. \cref{tab:ablations} reports the results. To measure whether appearances are assigned correctly, we use the class predicted by our fine-tuned detector, which distinguishes the four skins: when a player is rendered in a generated frame, the predicted class matches the skin of the corresponding character in 95.55\% of cases (see also the supplementary video). The detection rate and the visual quality metrics are comparable to the model trained with a single skin, so supporting several appearances does not reduce the consistency of player rendering. Since the multi-skin dataset is smaller than the dataset of the main experiments, the result also indicates that our model is able to converge even with less training data.

\begin{table}[t]
  \centering
\caption{\textbf{Ablations.}
The first block of methods tests the influence of a progressively more expressive state on the results. \textit{Varied skin} shows characters with different appearances. In \textit{1\,vs.\,1}, evaluation is restricted to the unmasked target player. Run on 30\,s rollouts. MILO values scaled by $10^{3}$. Best results are in \textbf{bold} (only first block).}
  \label{tab:ablations}
  \resizebox{\columnwidth}{!}{%
\begin{tabular}{@{}lccccc@{}}
  \toprule
  Method
    & LPIPS\,$\downarrow$
    & MILO\,$\downarrow$
    & FVD$_{32}$\,$\downarrow$
    & Recall [\%]\,$\uparrow$
    & Prec. [\%]\,$\uparrow$ \\
  \midrule



%
   Baseline                 & 0.344 & 13.05 & 299.8 & 28.3 & 40.3 \\
  Position only            & 0.315 & 11.58  & 279.9 & 36.4 & 44.9 \\
  Full State               & 0.280 & 10.57 & 223.1 & 28.4 & 38.6 \\
  + FOV-guided cross att.        & 0.266 & 10.48  & 152.9 & 66.5 & 66.2 \\
  + Dense player mask       & \textbf{0.249} & \textbf{9.53}
                           & \textbf{137.5}
                           & \textbf{71.1} & \textbf{70.1} \\
  \midrule
  Varied Skins             & 0.284 & 10.05  & 135.1 & 59.1 & 60.5 \\
  \midrule
  1 vs.\ 1                 & 0.279 & 10.15 & 259.1 & 68.2 & 70.0 \\
  \bottomrule

  \end{tabular}%
  }
\end{table}

\paragraph{Rollout Duration.}
We analyze the influence of the rollout duration on the results and compare windows of 10\,seconds, a duration used for evaluation by most prior works \cite{po2026multigen,ye2026worldcast, savva2026solaris,WHAMNature,guo2025ctrl}, 30\,seconds, and an extreme rollout of 10\,minutes. \cref{tab:comparison} shows the results. Visual quality decreases for all methods as the rollout becomes longer. The decrease is expected, since autoregressive generation can drift over time, and any deviation from the GT frames lowers the full-reference metrics. Our method nevertheless achieves the best visual quality on all metrics in every rollout window. The detection metrics, on the other hand, highlight our method's advantage clearly: For rollouts of up to 30\,s, our method renders other characters accurately and recall and precision stay about 70\% even when increasing length. For the 10\,min window, both recall and precision still stay above 50\%, so other characters are still displayed at the correct position on the screen most of the time. In the same setting, the characters generated by the compared methods fade completely, with a recall of at most 2\% for each method.

\paragraph{Varied Number of Characters.}
Since our world model generates the view of only a single player from the FOV conditioning, the number of players can change without retraining. We test the property on a two-player variant of our game, using the world model of the main experiments trained with four players. We keep one pair of player slots and zero the tokens of the other players after the state encoder; attention to their tokens is masked. \cref{tab:ablations} and the supplementary material show that the model produces consistent results for this two-player setting, without any fine-tuning.


\section{Limitations and Discussion}

First, training \acronym requires synchronized gameplay videos, actions, and ground-truth states for each NPC, which are difficult to obtain from existing video datasets. We therefore use a custom-built environment for acquiring labeled data. Future work could explore fine-tuning pretrained world models with limited state-labeled gameplay data to reduce this dependency.

Second, our state model approximates character dynamics using motion priors and learned corrections, which can introduce errors over long rollouts. In particular, NPC positions and orientations are occasionally inaccurate, affecting the consistency of generated observations. Incorporating uncertainty-aware prediction or occasional visual feedback could help correct accumulated errors without requiring continuous rendering.

Finally, our model and evaluation focus on a single first-person shooter game with relatively simple graphics. Scaling to more diverse environments and agent behaviors remains an important direction for future work and a general challenge for world models due to the necessary intense computation and data overhead.

\section{Conclusion}
We present SWAM, a state-aware action world model that separates multi-agent state evolution from autoregressive visual generation. By combining lightweight action-driven state prediction with FOV conditioning, SWAM maintains consistent character dynamics without requiring rendered observations or explicit scene geometry. Experiments in a 3D first-person shooter demonstrate improved visual fidelity, character consistency, and long-horizon stability.
\section*{LLM Usage Statement}
We used LLMs to revise and improve the manuscript writing and develop code for part of the algorithm under supervision.
{
    \small
    \bibliographystyle{ieeenat_fullname}
    \bibliography{main}
}


\end{document}